\documentclass[journal,twoside,web]{ieeecolor}
\usepackage{tmi}
\usepackage[noadjust]{cite}
\usepackage{amsmath,amssymb,amsfonts}
\usepackage{algorithm}
\usepackage{graphicx}
\usepackage{textcomp}

\usepackage{bm} % bold math for vector
\usepackage{multirow}
\usepackage{tabularx}
\usepackage{booktabs}       % professional-quality tables
\usepackage{pifont}

\usepackage[noend]{algpseudocode} % pseudo code
\algnewcommand\algorithmicforeach{\textbf{for each}}
\algdef{S}[FOR]{ForEach}[1]{\algorithmicforeach\ #1\ \algorithmicdo}
\usepackage{fixltx2e} % subscript in text mode
\usepackage[table]{xcolor}
\definecolor{green}{rgb}{0.0, 0.5, 0.0}
\newcommand{\cmark}{\ding{52}}%
\newcommand{\xmark}{\ding{56}}%

\def\BibTeX{{\rm B\kern-.05em{\sc i\kern-.025em b}\kern-.08em
    T\kern-.1667em\lower.7ex\hbox{E}\kern-.125emX}}
\begin{document}
\title{A Chain of Diagnosis Framework for Accurate and Explainable Radiology Report Generation}

\author{Haibo Jin, Haoxuan Che, Sunan He, and Hao Chen, \IEEEmembership{Senior Member, IEEE}
\thanks{This work was supported by the Hong Kong Innovation and Technology Commission (Project No. MHP/002/22 and No. ITCPD/17-9), the Pneumoconiosis Compensation Fund Board, HKSARS (Project No. PCFB22EG01), National Key R\&D Program of China (Project No. 2023YFE0204000) and the Research Grants Council of the Hong Kong (Project Reference Number: T45-401/22-N).
\textit{(Corresponding author: Hao Chen.)}} 
\thanks{H. Jin, H. Che and S. He are with Department of Computer Science and Engineering, Hong Kong University of Science and Technology, Hong Kong, China (e-mail: \{hjinag,hche\}@cse.ust.hk, shebd@connect.ust.hk)}
\thanks{H. Chen is with the Department of Computer Science and Engineering, Department of Chemical and Biological Engineering and Division of Life Science, Hong Kong University of Science and Technology, Hong Kong, China (e-mail: jhc@cse.ust.hk)}
}

\maketitle

\begin{abstract}
Despite the progress of radiology report generation (RRG), existing works face two challenges: 1) The performances in clinical efficacy are unsatisfactory, especially for lesion attributes description; 2) the generated text lacks explainability, making it difficult for radiologists to trust the results. To address the challenges, we focus on a trustworthy RRG model, which not only generates accurate descriptions of abnormalities, but also provides basis of its predictions. To this end, we propose a framework named chain of diagnosis (CoD), which maintains a chain of diagnostic process for clinically accurate and explainable RRG. It first generates question-answer (QA) pairs via diagnostic conversation to extract key findings, then prompts a large language model with QA diagnoses for accurate generation. To enhance explainability, a diagnosis grounding module is designed to match QA diagnoses and generated sentences, where the diagnoses act as a reference. Moreover, a lesion grounding module is designed to locate abnormalities in the image, further improving the working efficiency of radiologists. To facilitate label-efficient training, we propose an omni-supervised learning strategy with clinical consistency to leverage various types of annotations from different datasets. Our efforts lead to 1) an omni-labeled RRG dataset with QA pairs and lesion boxes; 2) a evaluation tool for assessing the accuracy of reports in describing lesion location and severity; 3) extensive experiments to demonstrate the effectiveness of CoD, where it outperforms both specialist and generalist models consistently on two RRG benchmarks and shows promising explainability by accurately grounding generated sentences to QA diagnoses and images. 
\end{abstract}


\section{Introduction}
\label{sec1}

\IEEEPARstart{R}{adiology} report generation (RRG) aims to generate a medical report for a radiological image (e.g., chest X-ray), where the report describes detailed clinical findings of the given image. Therefore, RRG has the potential to relieve the workload of radiologists and it has attracted great attention from researchers in recent years. 

RRG is a challenging task as the model needs to first correctly understand the given radiological image, then summarize the findings into a paragraph that follows clinical standards. Therefore, the generated reports should be not only linguistically precise, but also clinically accurate. To this end, various methods have been proposed to improve model performance. For example, \cite{CSC20,YWG23} proposed memory mechanism to record similar patterns for feature enhancement; \cite{ZWX20,LWG21,LLC23,HZZ23} incorporated domain knowledge to assist report generation via knowledge graphs; \cite{JXX18,WHW22,YaP22,YHL21,WZW21} introduced multi-task learning for better feature representations. More recently, Jin et al.~\cite{JCL24} pointed out the insufficiency of existing models in clinical efficacy and proposed diagnosis-driven prompts to guide the model for generating diagnostically accurate reports. Liu et al.~\cite{LTC24} introduced large language models (LLMs) with vision-text alignment strategy and coarse-to-fine decoding, which obtains competitive performance in both clinical efficacy (CE) and natural language generation (NLG) metrics. Chen et al.~\cite{CBJ25} proposed the first region-guided CT report generation framework that enhances diagnostic accuracy by integrating local anatomical features with global context and Che et al.~\cite{CJG25} investigated report generation in federated learning for the first time.

Despite the success of the previous methods, two major challenges still exist. First of all, whether the generated reports describe abnormalities precisely remains underexplored. Most works investigated the diagnostic accuracy as CE, yet the attributes of abnormalities also play an important role in condition assessment and surgery planning. For example, a severe cardiomegaly and a mild one makes a big difference for treatment decision; multiple lesions in both lungs and a single lesion in the left would require different surgery plans. Therefore, it is essential to ensure the attributes of lesions are correctly described. Furthermore, existing models often lack explainability. When a medical image is given, a report is simply output without providing evidences for how the result is generated. This makes it difficult for radiologists to trust the automatically generated reports and they may have to spend extra time to verify the results. To summarize, the two challenges make the existing models untrustworthy for the applications in clinical practice.

Given these challenges, there is an urgent need for a framework that not only improves the clinical accuracy of generated reports, but also provides transparent diagnostic reasoning to enhance trustworthiness. To achieve this, we argue that the report generation process should mirror the diagnostic workflow of radiologists, where abnormalities are systematically identified, analyzed, and documented with supporting evidence. Notably, LLMs possess strong reasoning capabilities when guided with structured intermediate steps, as demonstrated by the success of chain-of-thought (CoT)~\cite{WWS22}. Inspired by CoT, we propose to decompose the complex radiology diagnosis task into explicit, interpretable reasoning steps to leverage the LLM’s potential for accurate and explainable report generation.

To this end, we propose a framework named chain of diagnosis (CoD), which maintains a diagnostic chain by mimicking the diagnostic workflow of radiologists. Specifically, given a radiological image, the framework first generates question-answer (QA) pairs via diagnostic conversation in a self-talk manner to extract key findings of the image, then the QA diagnoses are used as a prompt to guide a LLM for clinically accurate generation of reports. To enhance explainability, a diagnosis grounding module is designed to match the QA diagnoses and the generated sentences of the report, where the diagnoses act as a reference of the report. Moreover, a lesion grounding module is designed to locate the mentioned abnormalities in the image, further improving the working efficiency of radiologists. To facilitate label-efficient training, we also propose an omni-supervised learning strategy with clinical consistency to leverage various types of annotations from different datasets. Experiments on two RRG benchmarks demonstrate the superiority of our model, which outperforms the existing methods consistently in CE metrics, including disease diagnosis, location prediction, and severity estimation. Additionally, experiments on diagnosis and lesion grounding are conducted to show the promise of our model in explainability. We summarize our contributions as follows.

\begin{enumerate}
\item We propose a new RRG framework named CoD for generating clinically accurate reports. Different from prior works, CoD first extracts key findings from the image through a question-and-answer process, then incorporates the extracted findings into the text generation process to provide effective guidance. To our knowledge, this is the first work that brings the chain-of-thought strategy to the field of report generation.
\item A novel diagnosis grounding module is proposed to enhance the explainability of the model. By matching the generated sentences to the intermediate QA pairs, we provide the diagnostic basis of the generated reports. Additionally, a lesion grounding module is designed to locate the mentioned abnormalities in the image, which further provides interpretability by connecting location-related QA diagnoses to the image.
\item An omni-supervised learning strategy is proposed to facilitate model training with different types of annotations from different datasets. Pseudo-labels are estimated and filtered with thresholding and clinical consistency for the missing labels, maximizing the use of the data at hand.
\item An evaluation tool is developed to assess the report accuracy in describing lesion location and severity. Experiments on two RRG benchmarks demonstrate the SOTA performance of CoD in report generation and the experiments on diagnosis and lesion grounding show its effectiveness in enhancing model explainability.
\end{enumerate}   

\begin{figure*}[t]
\centering
  \includegraphics[width=1\linewidth]{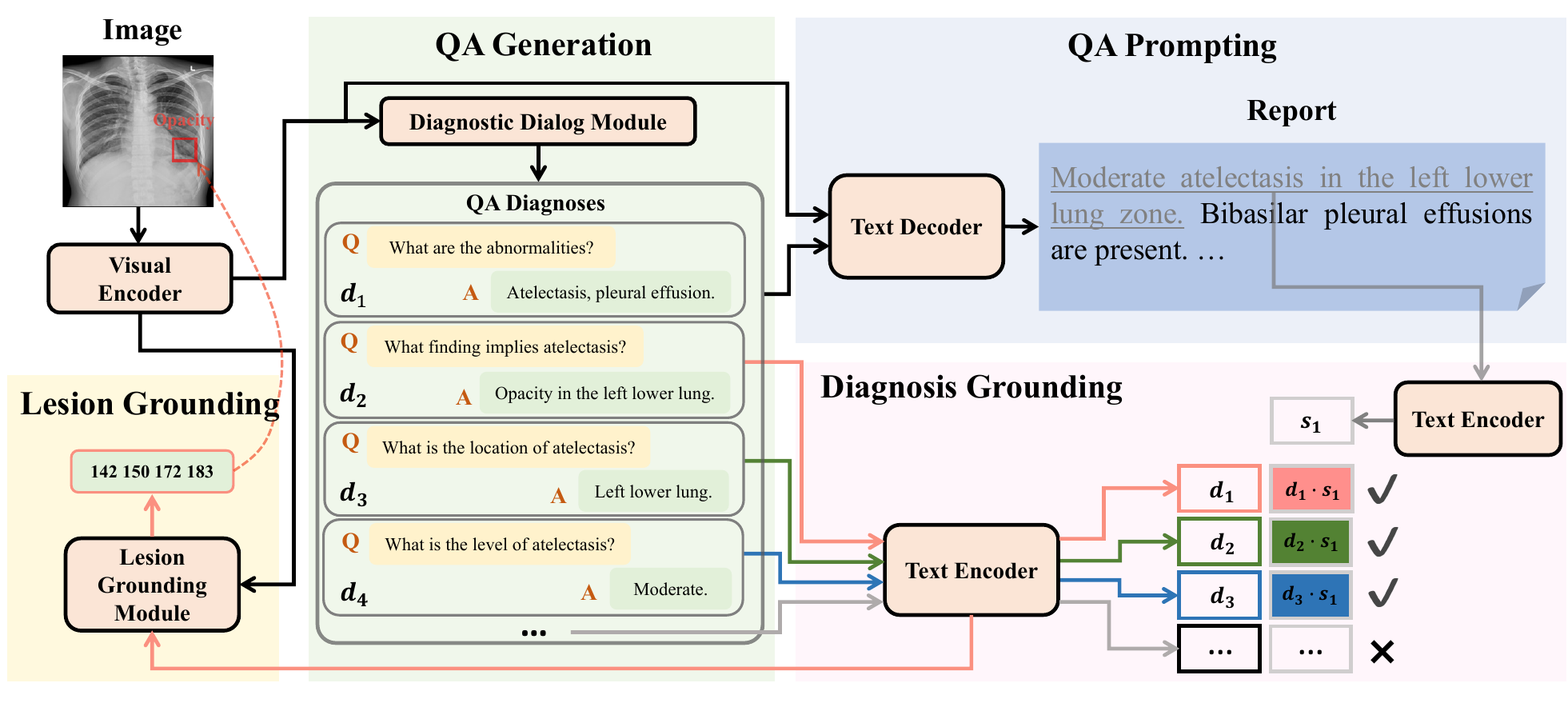}
\caption{The overall framework of CoD, which maintains a chain of the diagnostic process throughout the model. Given the medical image, QA diagnoses are first generated via the diagnostic dialog module, then used as the prompt to guide the text decoder for report generation. After that, the sentences in the report are matched with the QA diagnoses through the diagnosis grounding module. For the diagnosis related to lesion location, it will also be grounded in the image via the lesion grounding module.} 
\label{fig:framework}      
\end{figure*}

\section{Related Works}
\label{sec2}

\subsection{Radiology Report Generation}
\label{sec2.1}

Textual relevance and clinical efficacy are two essential objectives of RRG. To this end, various strategies have been proposed to improve model performance in terms of the two objectives. \cite{CSC20} and \cite{YWG23} imposed a memory mechanism to record past similar patterns for providing informative features to the text decoder. Guo et al.~\cite{GMX24} further combined memory module with prototype learning to avoid redundancy from visual features. Differently, \cite{ZWX20}, \cite{LWG21}, and \cite{HZZ23} constructed a static knowledge graph to denote the relationship between diseases and organs via graph neural networks, which allows for dedicated feature learning of the abnormalities. Later, Li et al.~\cite{LLC23} modified the static graph into a dynamic one, which consistently updates the graph by injecting new knowledge.  

Multi-task learning is a common technique to facilitate the representation learning of RRG, where auxiliary tasks are conducted besides report generation. Among the auxiliary tasks, disease classification is the most popular one as it helps the model to learn discriminative features~\cite{JXX18,WHW22,YaP22}. Similarly, weakly supervised contrastive learning was introduced by Yan et al.~\cite{YHL21} as an auxiliary task to learn a semantically meaningful space for better results. Additionally, image-text matching was explored~\cite{WHW22,WZW21,YaP22} to learn an aligned image-text representations in a fine-grained manner.

Due to the complex nature of report generation, divide-and-conquer is an effective way to ease the problem. RGRG~\cite{TMK23} designed a region-wise framework to generate sentences for each anatomical region separately. An object detector was trained to locate the predefined region first, then a text generator was used to output sentences related to the given region. Finally, the sentences were concatenated and post-processed to form a report. In contrast, Liu et al.~\cite{LHM19} chose to decode sentences hierarchically by first generating a topic vector for each sentence before word prediction and Jin et al.~\cite{JCL24} proposed to explicitly guide the text generator with the diagnostic results of a disease classifier. Our method also belongs to this category by decoupling the task into QA generation and QA-guided report generation. Compared to previous works~\cite{LHM19,JCL24}, our CoD uses free-form text to extract key information via self-talk conversation, which is more flexible and generalizable for various types of information.

\subsection{Chain-of-Thought}
\label{sec2.2}

CoT prompting~\cite{WWS22} has been shown to improve the reasoning capability of LLMs by utilizing a series of intermediate reasoning steps as prompts. Due to its effectiveness, various variants are designed to enhance the performance, such as tree-of-thought~\cite{YYZ24} and graph-of-thought prompting~\cite{BBK24}. Tree-of-thought explores multiple branches of reasoning to expand the searching space while graph-of-thought allows the model to explore interconnected ideas and dependencies by representing reasoning as a graph structure. Later, this idea was brought to vision-language tasks. For example, Multimodal-CoT~\cite{ZZL24} proposed a two-stage framework that separates rationale generation and answer inference for better reasoning capabilities based on multimodal information; Chen et al.~\cite{CZS24} addressed the knowledge-based visual reasoning task by converting key visual context into text context for prompting, termed visual chain-of-thought prompting. Similar to our approach, M4CXR~\cite{PKY24} employs CoT reasoning for chest X-ray interpretation to improve clinical accuracy. However, while their work primarily focuses on multi-task performance across report generation, visual grounding, and visual question answering, our framework specifically targets two critical aspects: (1) enhancing the accuracy of fine-grained attribute prediction and (2) addressing the explainability challenges in radiology report generation.

\subsection{Explainability in Medical Imaging}
\label{sec2.3}
Explainability in medical imaging has gained significant attention in recent years due to the increasing use of deep learning models in diagnostic processes. While these models have demonstrated remarkable accuracy in tasks such as tumor detection, segmentation, and disease classification, their ``black-box" nature has raised concerns about trust and adoption in clinical practice. Early works in this field focused on saliency maps and heatmaps, such as Grad-CAM~\cite{SCD20}, which highlights regions of interest in images that influence model decisions. However, these techniques often lack robustness and can be prone to highlighting irrelevant regions. More recent approaches include model-agnostic methods like LIME~\cite{RSG16} and SHAP~\cite{LuL22}, which aim to provide post-hoc explanations by approximating model behavior, as well as self-explaining models that generate interpretable outputs alongside predictions. Despite significant progress in explainable AI, most existing methods focus primarily on classification tasks, with few addressing the explainability challenges of more complex tasks like radiology report generation. MAIRA-2~\cite{BBC24} represents an early effort in this direction, proposing sentence-level grounding to image regions for easier verification. While our framework incorporates a similar lesion grounding module, we advance this approach by introducing dual grounding: (1) Diagnosis grounding, which explicitly connects each generated sentence to its corresponding intermediate QA-based diagnostic reasoning, and (2) Lesion grounding, which maps these diagnoses to their visual evidence in the input image.

\section{Method}
\label{sec3}

In this section, we first present the overall framework of CoD in Sec.~\ref{sec3.1}, then introduce the QA generation and prompting modules in Sec.~\ref{sec3.2} and the grounding modules in Sec.~\ref{sec3.3}. The proposed omni-supervised learning strategy is given in Sec.~\ref{sec3.4}. Finally, we give the details of the developed new CE metrics in Sec.~\ref{sec3.5}.

\subsection{Framework}
\label{sec3.1}

Fig.~\ref{fig:framework} shows the overall framework of CoD. Given the medical image, QA diagnoses are first generated via the diagnostic dialog module, then used as a prompt to guide the LLM decoder for report generation. After that, the sentences in the report are matched with the QA diagnoses through the diagnosis grounding module. For the diagnosis related to lesion location, it will also be grounded in the image via the lesion grounding module.

Formally, the visual feature extraction is denoted as
\begin{equation}
f_{\text{ve}}(I) = \bm{X} = \{\bm{x}_1, \bm{x}_2, ..., \bm{x}_S\},
\end{equation}
where $f_{\text{ve}}$ is visual encoder, $I$ is input image, $\bm{x}_i \in \mathbb{R}^{C}$ denotes the feature of an image patch, $S$ denotes the number of patches, and $C$ represents the dimension of feature. We denote each report as $R=\{ r_1, r_2, ..., r_T \}, r_i \in \mathbb{V}$, where each $r_i$ is a token, $T$ is the length of the report, and $\mathbb{V}$ represents the vocabulary. The process of decoding is formulated as
\begin{equation}
\label{eq_report_dec}
r_t = f_{\text{td}}(\bm{X}, D, r_1,...,r_{t-1}),
\end{equation}
where $f_{\text{td}}$ is text decoder, $r_t$ is the token to be predicted at time step $t$ and $D$ represents QA diagnoses. The details of QA diagnoses generation and prompting are given in Sec.~\ref{sec3.2}. Language modeling loss is used for report generation:
\begin{equation}
\label{eq_lm_loss}
\mathcal{L}_{\text{LM}} = -\sum_{t=1}^T \log p(r_t|\bm{X},D,r_1,...,r_{t-1}).
\end{equation}

During diagnosis grounding, we aim to find matched QA diagnoses for a given sentence $s_i$. We denote this process as $\hat{D}_i = \{ d_j \in D \mid S(s_i, d_j) > \gamma \}$, where $\hat{D}_i$ is the matched set of diagnoses, $D=\{ d_1,...,d_n \}$ is the set of QA diagnoses generated from the image, $S(s_i, d_j)$ represents the similarity score between the sentence $s_i$ and the diagnosis $d_j$, and $\gamma$ is the threshold. The training details of diagnosis grounding are given in Sec.~\ref{sec3.3}.

When a QA diagnosis involves location-related content, we treat it as a text input to visually ground it in the image. We name this process as lesion grounding and its process can be represented as
\begin{equation}
B = f_{\text{lg}}(\bm{X}, d_{\text{loc}}),
\end{equation}
where $f_{\text{lg}}$ is the lesion grounding module, $d_{\text{loc}}$ is the QA diagnosis related to lesion location, and $B=(x_1, y_1, x_2, y_2)$ is the predicted bounding box, defined by its top-left $(x_1, y_1)$ and bottom-right $(x_2, y_2)$ coordinates. The detailed architecture of the module can be seen in Sec.~\ref{sec3.3}.

\subsection{QA Generation and Prompting}
\label{sec3.2}

For the task of report generation, it is crucial to ensure that generated texts align with the medical image. This is because the report of a medical image must not only offer a thorough summary but also capture its clinical relevance. If the generated report fails to capture essential details of abnormalities, it could lead to incorrect conclusions about the examination, potentially resulting in severe consequences. Previous works have proposed various solutions to address this issue, yet they simply treat diagnostic accuracy as clinical efficacy while ignoring the attributes of abnormalities. We argue that these attributes play an important role in condition assessment and surgery planning. For example, a severe cardiomegaly and a mild one makes a big difference for treatment decision; multiple lesions in both lungs and a single lesion in the left would require different surgery plans. Therefore, it is essential to ensure the attributes of lesions are correctly described.

To this end, we propose QA generation and prompting, a method that queries key information from the image via dialog, then prompts the text decoder with these information for clinically accurate report generation. Specifically, our text decoder is trained to ask questions and answer it by itself sequentially, where the QA pairs are curated from existing datasets such as MIMIC-Diff-VQA~\cite{HGA23}. Formally, we have
\begin{equation}
d_t = f_{\text{td}}(\bm{X}, d_1,...,d_{t-1}),
\end{equation}
where $f_{\text{td}}$ is the same text decoder as report generation and $d_t$ is the $t$-th QA diagnosis that consists of a QA pair. Four types of questions are involved, namely (1) presence of disease, (2) its location, (3) severity, and (4) underlying causal relation. The same language modeling loss as Eq.~\ref{eq_lm_loss} is used here for the training of diagnostic dialog. After the dialog $D$ is generated, it will be used as a prompt to guide the report generation process, as shown in Eq.~\ref{eq_report_dec}.

\subsection{Diagnosis and Lesion Grounding}
\label{sec3.3}

In RRG, most existing works focus solely on improving model performance while largely neglecting the critical issue of explainability. Given that RRG systems are designed to assist rather than replace radiologists, establishing trust in model predictions is essential, which requires moving beyond "black-box" systems to the ones capable of transparent reasoning. However, achieving explainability in RRG remains an open challenge, as it is inherently more complex than in classification tasks. To address this, we propose a novel approach that enhances explainability through dual grounding: (1) \textbf{diagnosis grounding}, which explicitly links each generated sentence to its corresponding intermediate QA-based diagnostic reasoning, and (2) \textbf{lesion grounding}, which connects these diagnoses to their visual evidence in the input image. This two-level grounding provides radiologists with a verifiable basis for every generated report, enabling them to efficiently verify consistency between the report and its supporting evidence. Crucially, if any discrepancy is identified, radiologists can directly correct the specific QA diagnosis, and the system will automatically update the report accordingly, simplifying the verification process while maintaining clinical workflow efficiency.

\begin{figure}[t]
\centering
  \includegraphics[width=0.8\linewidth]{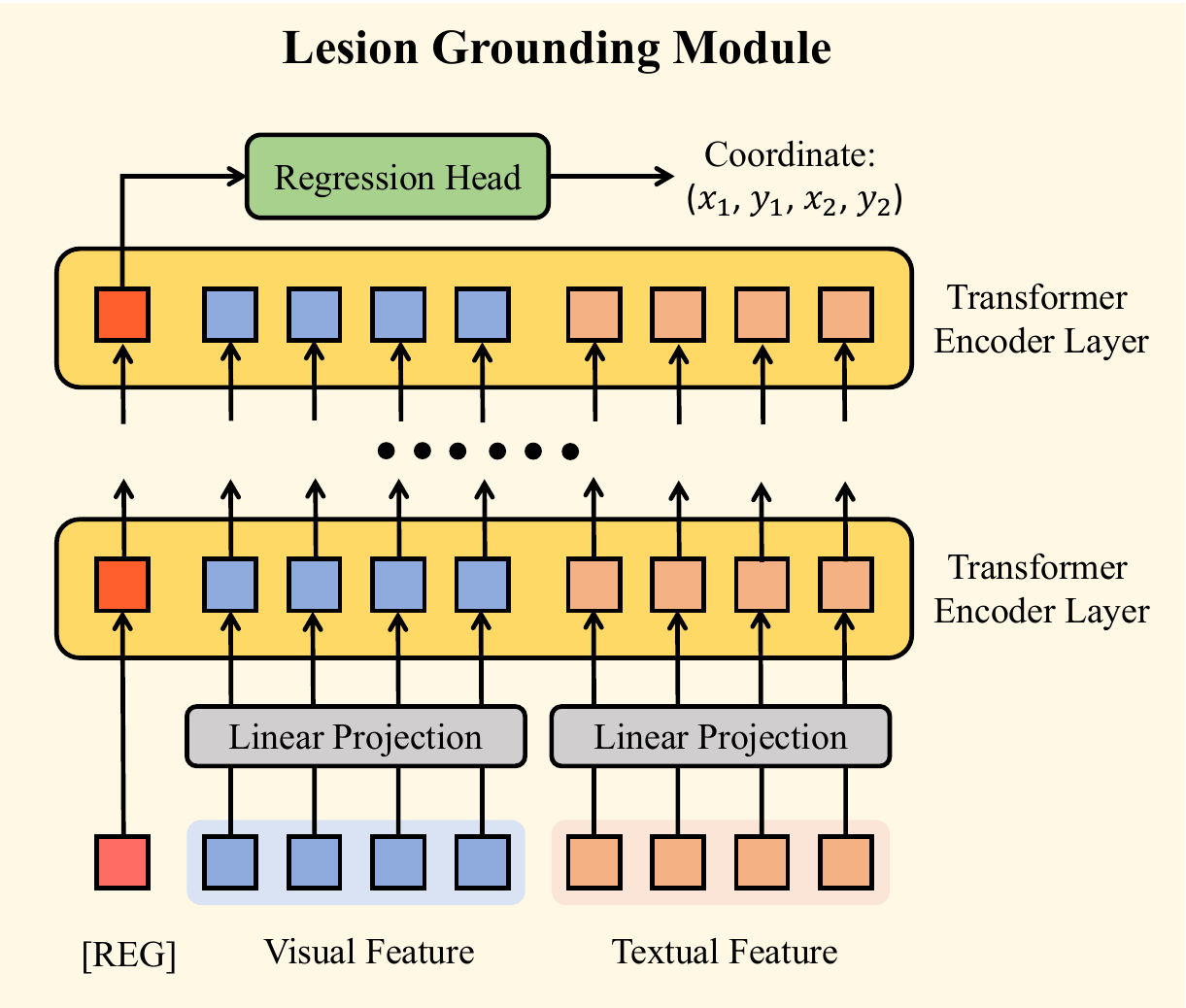}
\caption{Architecture of lesion grounding module. The visual and textual features first go through the projection layer for dimension reduction, then concatenated with a special token \texttt{[REG]} to form the multimodal feature. After six transformer encoder layers , the embedding of \texttt{[REG]} is extracted and input to the regression head for predicting the coordinate of lesion.} 
\label{fig:lesion_grounding}      
\end{figure}

\begin{figure*}[t]
\centering
  \includegraphics[width=0.83\linewidth]{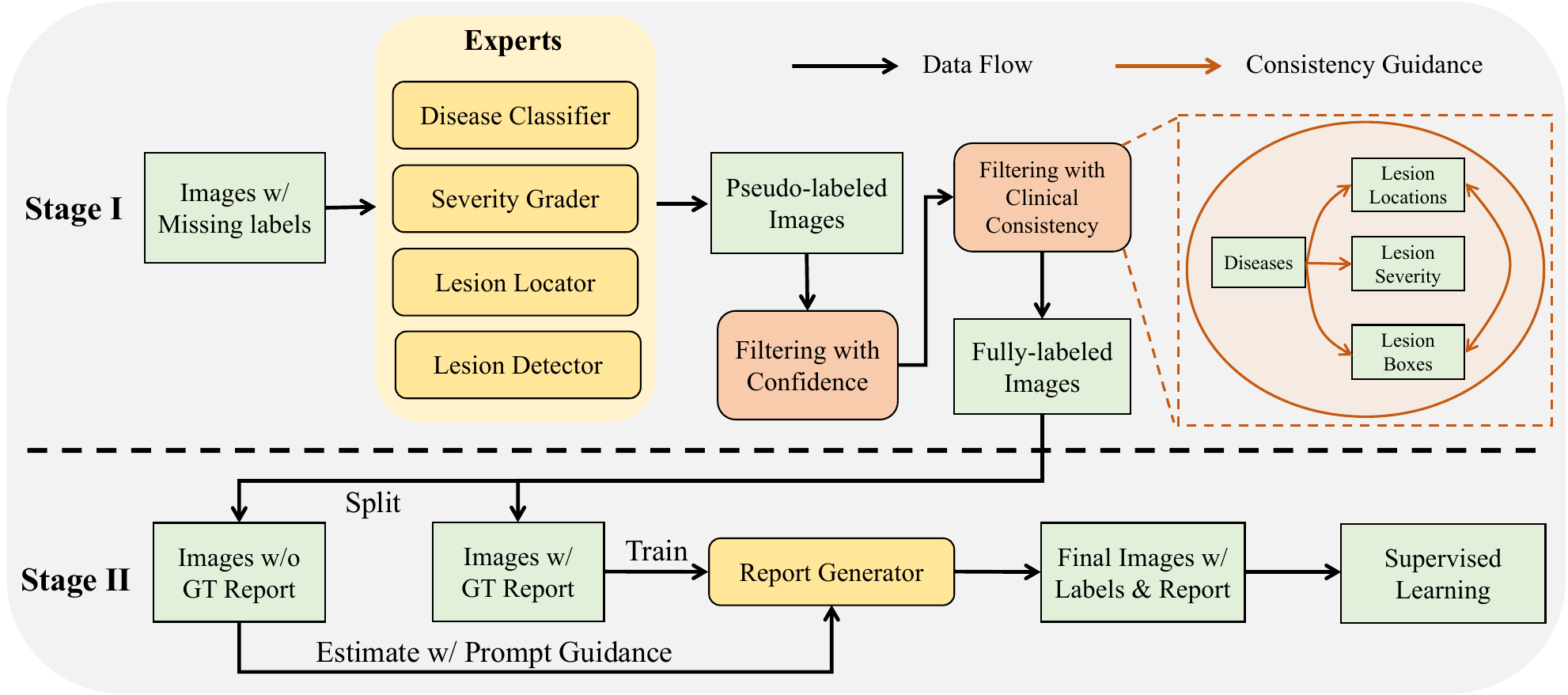}
\caption{Our omni-supervised learning strategy. It estimates pseudo-labels in the first stage and pseudo reports in the second stage.} 
\label{fig:omni}      
\end{figure*}

\noindent
\textbf{Diagnosis grounding.}
We formulate diagnosis grounding as a matching task between report sentences and QA diagnoses. Since a sentence often cover abnormality presence, location, and severity at the same time, it is possible a sentence is matched to more than one diagnosis and vice versa. To facilitate training for the multi-positive cases, we adopt the Multi-Positive NCE loss~\cite{LKS22} to force a subject to be close to all the positive pairs while keeping distance to negative pairs. We use $\mathbb{P}_i$ to denote the set of positive pairs of $i$-th subject and $\mathbb{N}_i$ for negative pairs. Following CLIP~\cite{RKH21}, the contrastive loss is imposed to both sides for stable convergence. Thus, the training loss of the diagnosis grounding is formulated as:
\begin{equation}
\mathcal{L}_{\text{DG}} = \frac{1}{2N} \sum_{i=1}^{N} ( \ell_i^{\text{d2s}} + \ell_i^{\text{s2d}} ),
\end{equation}
\begin{equation}
\ell_i^{\text{d2s}} = \mathbb{E}_{j \in \mathbb{P}_i} \left[  \log \frac{e(d_i,s_j)}{e(d_i,s_j)+\sum_{k \in \mathbb{N}_i} e(d_i,s_k)} \right],
\end{equation}
\begin{equation}
\ell_i^{\text{s2d}} = \mathbb{E}_{j \in \mathbb{P}_i} \left[  \log \frac{e(s_i,d_j)}{e(s_i,d_j)+\sum_{k \in \mathbb{N}_i} e(s_i,d_k)} \right],
\end{equation}
where $e(d_i, s_j)=\exp(\text{sim}(f_{\text{te}}(d_i), f_{\text{te}}(s_j)) / \tau)$, $ f_{\text{te}}$ is the text encoder, sim  indicates the cosine similarity, and $\tau$ is a temperature parameter. Note that the training of this module is independent to the overall framework because it does not require images for training.

\noindent
\textbf{Lesion grounding.}
Fig.~\ref{fig:lesion_grounding} shows the architecture of lesion grounding module. We adopt the model from TransVG\cite{DYC21} while other alternatives are feasible (e.g., MedRPG~\cite{CZT23}). After the visual feature $\bm{X} \in \mathbb{R}^{C_v \times S_v}$ and textual feature $\bm{Z} \in \mathbb{R}^{C_t \times S_t}$ are extracted, they first go through a projection layer to project the dimension to $C_p$, then the two features are concatenated with a special token \texttt{[REG]} to form the multimodal feature $\bm{M} \in \mathbb{R}^{C_p \times (S_v+S_t+1)}$, where 1 is the token length of \texttt{[REG]}, and $S_v$ and $S_t$ are the lengths of visual and textual tokens, respectively. Six transformer encoder layers are then used to conduct information exchange for the multimodal feature $\bm{M}$. Finally, the embedding of \texttt{[REG]} is input to the regression head for outputing the coordinate of target bounding box.

\subsection{Omni Learning with Clinical Consistency}
\label{sec3.4}

When the data is all available, our model can be easily trained following the framework in Fig.~\ref{fig:framework}. However, it is not easy to obtain a dataset with full annotations that cover various diagnostic information. In contrast, it is more common that different types of annotations are spread across different datasets. For example, MIMIC-CXR~\cite{JPG19} contains radiology reports; VinDR-CXR~\cite{NLL20} contains classification labels and bounding boxes; MIMIC-Diff-VQA~\cite{HGA23} provides QA pairs. To facilitate model training under realistic scenarios, we propose to train our model with an omni-supervised learning strategy such that different types of annotations are maximally utilized. 

\begin{figure}[t]
\centering
  \includegraphics[width=1\linewidth]{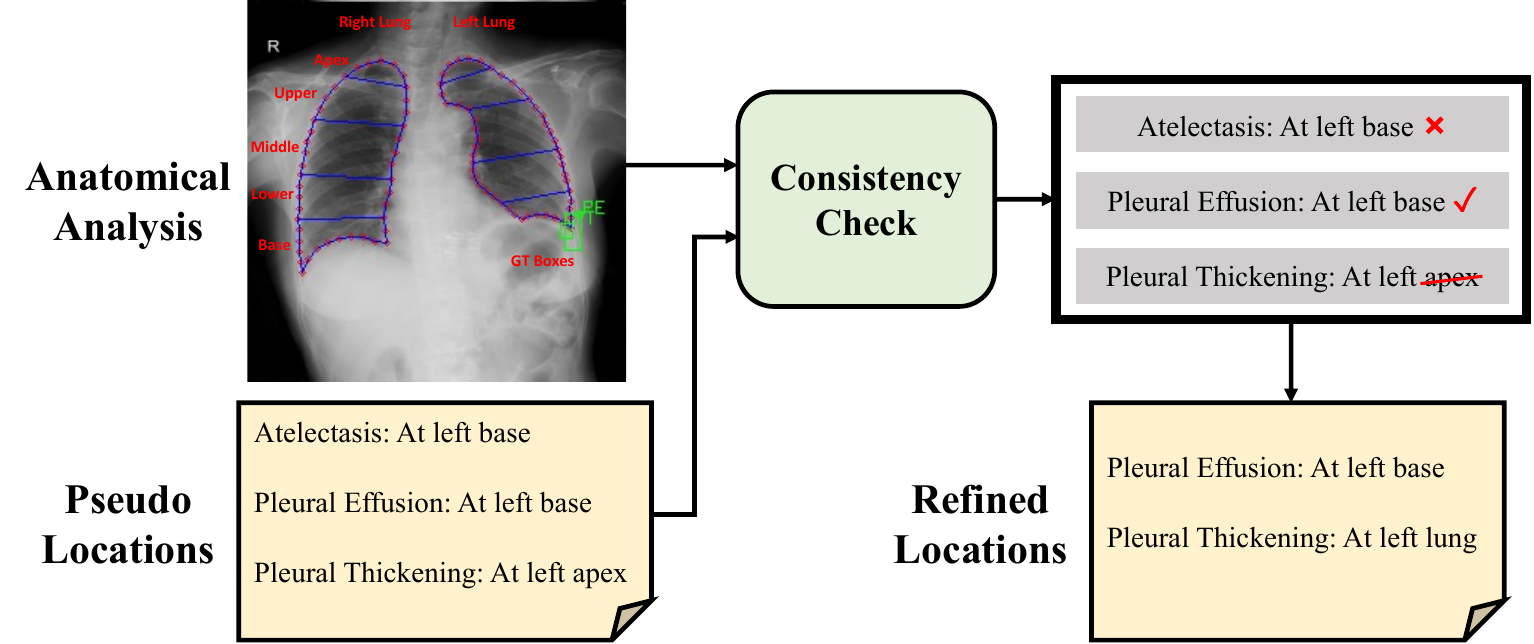}
\caption{An example of clinical consistency that refines the pseudo locations with GT bounding boxes via anatomical analysis.} 
\label{fig:consistency}      
\end{figure}

Fig.~\ref{fig:omni} shows the proposed omni-supervised learning strategy, which consists of two stages. In the \textbf{first stage}, we first train a set of experts for different types of annotations, including disease classification, severity grading, lesion location prediction, and lesion detection. Given a sample that misses certain types of annotations, we estimate pseudo-labels for these missed types using the expert models. Due to the potential noise in pseudo-labels~\cite{JLS21,JCC23}, it is necessary to filter unreliable pseudo-information so that it does not cause performance degradation. To this end, we first use confidence score to filter out the pseudo-labels below the threshold, then utilize the consistency across different types of annotations to further filter out unreliable pseudo-labels.

The right part of Fig.~\ref{fig:omni} gives the dependency relationship of such a clinical consistency. As we can see, the ground-truth (GT) label of disease classification could be used to guide the pseudo-label of lesion location, severity, and bounding boxes. In other words, if a disease is not present in the classification label, then the relevant prediction in other types will be removed. Moreover, the consistency between lesion location and bounding box will be checked. Fig.~\ref{fig:consistency} gives an example of the consistency check between predicted pseudo locations and GT bounding boxes. Specifically, we separately train a lung landmark detection model to identify the contour of both lungs based on the model from~\cite{JCC24}. Following the method in \cite{ZLL14} and the location annotation in \cite{HGA23}, we further divide each lung into five subregions vertically, namely 1) apex, 2) upper, 3) middle, 4) lower, and 5) base. After that, we conduct anatomical analysis by calculating which subregions the GT boxes fall in, then refine the pseudo locations by removing inconsistent location predictions based on the anatomical analysis. Note that we do not simply filter out the unreliable samples as in general semi-supervised learning~\cite{SBC20}; alternatively, we only remove the inconsistent labels of the sample while keeping the correct ones. By doing so, we not only retain sufficient number of training samples, but also make the model adapt to incomplete QA prompts. Similarly, such a consistency check can also be used for pseudo bounding boxes and GT locations.

\begin{algorithm}[t]
\caption{Our omni-supervised learning algorithm}
\begin{algorithmic}[1]
\Require All training data, with missing labels/reports.
\Ensure Fully annotated training data.
\State \textit{\textbf{Stage I}}
\State Train expert models using available labels.
\For{each training sample} 
\If {the sample has missing label}
\State Estimate pseudo-labels using expert models.
\State Filter out unreliable pseudo-labels using confidence score and clinical consistency.
\EndIf
\EndFor
\State \textit{\textbf{Stage II}}
\State Split the data into two parts: $D_1$ (w/ GT report) and $D_2$ (w/o GT report).
\State Train a report generator on $D_1$ using QA prompting.
\For{each sample in $D_2$} 
\State Estimate report with QA diagnoses as guidance.
\EndFor
\State Train a final model in supervised learning w/ above data.
\end{algorithmic}
\label{alg1}
\end{algorithm}

Inspired by the observation that diagnostic prompts can significantly improve the diagnostic accuracy when generating reports~\cite{JCL24}, we estimate pseudo reports for those missed ones with the guidance of QA diagnoses. Therefore, in the \textbf{second stage}, we split the output data from stage one into two groups, namely the samples with GT reports and the ones without. We then use the data with GT reports to train a report generator based on QA prompting, such that it could be used to estimate pseudo reports for those missed data. By doing this, all the missing labels and reports are filled and our final model could be trained in supervised learning, as described in Fig.~\ref{fig:framework}. The overall training pipeline is given in Alg.~\ref{alg1}.  

\subsection{Extended CE Metrics}
\label{sec3.5}

For RRG, evaluation on clinical efficacy is an important supplement to textual similarity as the former is more clinically relevant. R2Gen is one of the pioneering works that proposes to evaluate the clinical efficacy by testing the diagnostic accuracy of model and many follow-up works use the same setting for evaluation. However, we argue that the attributes of abnormalities also play an important role in clinical practice (e.g., condition assessment and surgery planning), which were neglected in the previous works. To make the evaluation more comprehensive, we develop a new evaluation tool to assess the accuracy in predicting lesion location and severity. Given a generated report, our tool could convert it into predefined classes in terms of lesion location and severity. Inspired by CheXbert~\cite{SJR20}, we propose AttriBert, a text classifier that uses Bert as the text encoder for text classification. To train it, we processed the annotation from MIMIC-Diff-VQA~\cite{HGA23} by selecting labels related to lesion location and severity. We categorized lesion severity into three classes, namely mild, moderate, and severe. For location, we decouple it into horizontal and vertical directions, where the horizontal direction has three categories (i.e., (1) left and right lung, (2) only left, and (3) only right) and the vertical has five classes (i.e., (1) apex, (2) upper, (3) middle, (4) lower, and (5) base). To improve generalization, we further included IU X-Ray~\cite{DKR16} by processing it in similar way, where its location and severity labels were obtained by querying Qwen-max~\cite{BBC23}. Finally, we curated 274,958 training samples and 3,858 test samples for AttriBert. Table~\ref{tab:results_attribert} shows the test results of our text classifier, which achieves 0.951, 0.954, and 0.973 accuracy on horizontal location, vertical location, and severity, respectively. We believe its performance is sufficient for evaluating the attribute accuracy of a given report and it will be used for testing the extended CE metrics in this paper.  

\begin{table}[t]
\centering
\newcolumntype{C}{>{\centering\arraybackslash}X}
\caption{Test accuracy of AttriBert on three new CE metrics.}
\begin{tabularx}{\linewidth}{lCCC}
\toprule	 
\textbf{Method} & Loc\textsubscript{h} & Loc\textsubscript{v} & Sev \\ 
\midrule
AttriBert & 0.951 & 0.954 & 0.973 \\
\bottomrule
\end{tabularx}
\label{tab:results_attribert}
\end{table}

\begin{table*}[t]
\centering
\scriptsize
\newcolumntype{C}{>{\centering\arraybackslash}X}
\caption{Comparison with SOTA RRG models (i.e., specialists) and foundation models (i.e., generalists) on IU X-Ray and MIMIC-CXR data. $*$ represents a generalist. The best results are in \textbf{bold} and the second best are \underline{underlined}.}
\begin{tabularx}{\linewidth}{lllCCCCCCCCC}
\toprule	 
\multirow{2}{*}{\textbf{Dataset}} & \multirow{2}{*}{\textbf{Model}} & \multirow{2}{*}{\textbf{Year}} & \multicolumn{6}{c}{\textbf{CE Metrics}} & \multicolumn{3}{c}{\textbf{NLG Metrics}}\\ \cmidrule(r){4-9} \cmidrule(r){10-12}
 & & &Loc\textsubscript{h} & Loc\textsubscript{v} & Sev & F1 & Rec & Pre & B1 & B4 & R \\
\midrule
\multirow{8}{*}{\textbf{IU X-Ray}} & R2Gen & 2020 & 0.292 & 0.495 & 0.688  & 0.136 & 0.136  & 0.141 & 0.325 & 0.059 & 0.253 \\
& CVT2Dis. & 2022 & \underline{0.346} & 0.476 & 0.702  & 0.168  & 0.172  & 0.174 & 0.383 & 0.082 & 0.277 \\
& M2KT & 2023 & 0.324 & 0.480 & 0.709  & 0.145  & 0.145  & 0.153 & 0.371 & 0.078 & 0.261 \\
& DCL & 2023 & 0.299 & \underline{0.495} & 0.695  & 0.162  & 0.167  & 0.168 & 0.354 & 0.074 & 0.267 \\
& RGRG & 2023 & 0.334 & 0.474 & 0.695  & 0.180  & 0.187  & 0.183 & 0.266 & 0.063 & 0.180 \\
& R2GenGPT & 2023 & 0.315 & 0.445 & 0.702  & 0.184  & 0.184  & 0.196 & 0.330 & 0.068 & 0.233 \\
& PromptMRG & 2024 & 0.341 & 0.485 & \underline{0.712}  & \underline{0.211} & \underline{0.229} & \underline{0.213} & \underline{0.401} & \textbf{0.098} & \underline{0.281} \\
\cmidrule(r){2-12}
& \textbf{CoD (Ours)} & - & \textbf{0.404} & \textbf{0.514} & \textbf{0.732} & \textbf{0.219} & \textbf{0.234}  & \textbf{0.218} & \textbf{0.403} & \underline{0.091} & \textbf{0.288} \\
\midrule
\multirow{21}{*}{\textbf{MIMIC}} 
& R2Gen & 2020 & 0.329 & 0.400 & 0.715 & 0.276 & 0.273 & 0.333 & 0.353 & 0.103 & 0.277 \\
& M2TR & 2021 & - & - & - & 0.308 & 0.428 & 0.240 & 0.378 & 0.107 & 0.272 \\
& MKSG & 2022 & - & - & -  & 0.371 & 0.348  & 0.458 & 0.363 & 0.115 & 0.284 \\
& CliBert & 2022 & - & - & -  & 0.415 & 0.435  & 0.397 & 0.383 & 0.106 & 0.275 \\
& CVT2Dis. & 2022 & 0.423 & 0.423 & 0.717 & 0.384 & 0.412 & 0.356 & 0.392 & 0.124 & 0.285 \\
& M2KT & 2023 & 0.404 & 0.461 & \underline{0.722} & 0.352 & 0.339 & 0.420 & 0.386 & 0.111 & 0.274 \\
& METrans. & 2023 & - & - & -  & 0.311  & 0.309  & 0.364 & 0.386 & 0.124 & \textbf{0.291} \\
& KiUT & 2023 & - & - & -  & 0.321  & 0.318  & 0.371 & 0.393 & 0.113 & 0.285 \\
& DCL & 2023 & 0.427 & 0.451 & \underline{0.722}  & 0.373  & 0.352  & 0.471 & - & 0.109 & 0.284 \\
& RGRG & 2023 & 0.405 & \underline{0.469} & 0.699  & 0.447  & 0.475  & 0.461 & 0.373 & 0.126 & 0.264 \\
& R2GenGPT & 2023 & 0.390 & 0.443 & 0.710  & 0.411  & 0.429  & 0.453 & 0.389 & 0.107 & 0.259 \\
& PromptMRG & 2024 & \underline{0.431} & 0.438 & 0.714  & \underline{0.476}  & \underline{0.509} & \textbf{0.501} & 0.398 & 0.112 & 0.268 \\
& I3+C2FD & 2024 & - & - & - & 0.473 & 0.482 & 0.465 & \underline{0.402} & \underline{0.128} & \textbf{0.291} \\
\cmidrule(r){2-12}
& Med-Flamingo$^*$ & 2023 & 0.344 & 0.193 & 0.667 & 0.144 & 0.128 & 0.207 & 0.221 & 0.035 & 0.197 \\
& RadFM$^*$ & 2023 & 0.305 & 0.332 & 0.694 & 0.184 & 0.184 & 0.212 & 0.187 & 0.044 & 0.199 \\
& InternVL$^*$ & 2024 & 0.346 & 0.310 & \underline{0.722} & 0.155 & 0.141 & 0.215 & 0.236 & 0.026 & 0.185 \\
& LLaVA-Med$^*$ & 2024 & 0.313 & 0.161 & 0.657 & 0.154 & 0.143 & 0.208 & 0.220 & 0.021 & 0.156 \\
& MedDr$^*$ & 2024 & 0.361 & 0.429 & 0.708 & 0.306 & 0.289 & 0.377 & 0.238 & 0.049 & 0.219 \\
\cmidrule(r){2-12}
& \textbf{CoD (Ours)} & - & \textbf{0.514} & \textbf{0.495} & \textbf{0.749}  & \textbf{0.479} & \textbf{0.521} & \underline{0.487} & \textbf{0.412} & \textbf{0.129} & \underline{0.286} \\
\bottomrule
\end{tabularx}
\label{tab:results_sota}
\end{table*}

\section{Experiments}

\subsection{Datasets}

\textbf{MIMIC-CXR}~\cite{JPG19} is the largest dataset for radiology report generation based on chest X-ray images. Following previous works, we split the data into training (270,790 samples), validation (2,130 samples), and test (3,858 samples) sets. 

\textbf{IU X-Ray}~\cite{DKR16} is another dataset for report generation. In this work, we conduct evaluation on the whole set of IU X-Ray, following the setting in \cite{JCL24}. There are 4,168 test samples in total, including both frontal and lateral view of CXRs.

\textbf{Medical-Diff-VQA}~\cite{HGA23} is designed for difference visual question answering task, building upon MIMIC-CXR. We processed the data by extracting QA pairs related to diagnosis, lesion location, and severity, then associated the pairs to the matched images in MIMIC-CXR. After our processing, there are 58.7\%, 35.3\%, and 22.3\% positive images in MIMIC-CXR are attached with attributes on horizontal location, vertical location, and disease severity, respectively. 

\textbf{RSNA}~\cite{RSN18} provides pneumonia detection data of CXR. We selected the positive cases (5,659 samples) to be the test set of our lesion grounding task.

\textbf{VinDR-CXR}~\cite{NLL20} consists of CXRs with bounding boxes for multiple diseases. However, some diseases contain only a few samples and some are beyond the scope of our training data. Thus, we finally selected two diseases (pleural effusion and lesion) and the processed number of images is 3,706. This data is also used as the test set of lesion grounding. 

\textbf{Private} data is also utilized in this work. We collected 15,555 frontal CXR images, with bounding box annotations on five diseases, including pneumonia, pleural effusion, pneumothorax, lesion, and fracture. Each image was labeled and verified by two radiologists under the assistance of reports. We use this private dataset as the training data to provide precise location information such that our model is able to localize lesions in the image.

\textbf{DG} is a dataset curated by us for diagnosis grounding, based on MIMIC-CXR and Medical-Diff-VQA. Specifically, we matched the extracted QA pairs of Medical-Diff-VQA to the sentences of MIMIC-CXR reports via keyword matching, which yields many noisy pairs of diagnosis and report sentence. We then utilized Qwen-max to help filter out incorrect matches by giving carefully designed instructions. Finally, we randomly split these data into training and test set, and manually checked the test set to ensure correctness. The processed data contains 302,077 and 15,836 pairs of examples for training and testing, respectively.

\subsection{Experimental Settings}

\noindent
\textbf{Evaluation metrics.}
For RRG, we use two types of metrics, namely NLG and CE. The NLG metrics include BLEU~\cite{PRW02} and ROUGE-L~\cite{Lin04}. BLEU is a precision-based metric that measures the n-gram overlap between the generated and reference text while ROUGE-L emphasizes the recall of the generated text by considering the longest common subsequence. The CE metrics include the commonly used precision, recall, and F1, as well as the newly developed attribute accuracy of horizontal location, vertical location, and severity. We use CheXbert\footnote{Although CheXbert is not conventionally used for IU X-Ray, our rigorous evaluation using Qwen-max reveals superior label extraction accuracy on IU X-Ray compared to MIMIC-CXR (0.967 vs. 0.944). This enhanced performance likely stems from IU X-Ray reports' greater conciseness and lower lexical diversity, which appear to facilitate more accurate label identification.}~\cite{SJR20} to convert reports into 14 disease classification results and use the developed AttriBert to convert reports into attribute categories, covering three horizontal locations, five vertical locations, and three severity levels. For lesion grounding, we adopt the Acc@0.5 from general grounding task~\cite{DYC21}, which considers the predicted region correct if its IoU is at least 0.5 with the GT bounding box. For diagnosis grounding, we calculate Top-N accuracy from 16 candidates, where N ranges from 1 to 3.

\begin{table*}[t]
\centering
\footnotesize
\newcolumntype{C}{>{\centering\arraybackslash}X}
\caption{Test precision and recall of lesion grounding on RSNA and VinDR-CXR. The best results are in \textbf{bold}.}
\begin{tabularx}{\linewidth}{lCCCCCCll}
\toprule	 
 & \multicolumn{2}{c}{\textbf{RSNA}} & \multicolumn{4}{c}{\textbf{VinDR-CXR}} & \\ \cmidrule(r){2-3} \cmidrule(r){4-7}
\textbf{Method} & \multicolumn{2}{c}{Pneumonia} & \multicolumn{2}{c}{Lesion} & \multicolumn{2}{c}{Pleural Effusion} & \multicolumn{2}{c}{\textbf{Average}} \\ 
\cmidrule(r){2-3} \cmidrule(r){4-5} \cmidrule(r){6-7} \cmidrule(r){8-9}
 & Prec. & Rec. & Prec. & Rec. & Prec. & Rec. & Prec. & Rec. \\ 
\midrule
Baseline & 0.465  & 0.477 & 0.106  & 0.126 & 0.302  & 0.382 & 0.336 & 0.344 \\	
+Omni & 0.498  & 0.502 & 0.114  & 0.129 & 0.291  & 0.368 & 0.354 (+5.4\%) & 0.362 (+5.2\%) \\
+Omni +CC & 0.546  & 0.576 & 0.146  & 0.188 & 0.354  & 0.464 & \textbf{0.401} (+19.3\%) & \textbf{0.423} (+23.0\%) \\
\bottomrule
\end{tabularx}
\label{tab:results_vg}
\end{table*}

\begin{table}[t]
\centering
\scriptsize
\newcolumntype{C}{>{\centering\arraybackslash}X}
\caption{Test results of diagnosis grounding in top-N accuracy, denoted with mean and standard deviation.}
\begin{tabularx}{\linewidth}{lCCC}
\toprule	 
\textbf{Method} & Top-1 & Top-2 & Top-3 \\ 
\midrule
Keyword & 0.129\tiny{±0.002} & 0.231\tiny{±0.001} & 0.309\tiny{±0.002} \\
Bert & 0.114\tiny{±0.002} & 0.194\tiny{±0.003} & 0.260\tiny{±0.002} \\
CLIP & 0.202\tiny{±0.002} & 0.330\tiny{±0.002} & 0.411\tiny{±0.002} \\
CXR-CLIP & 0.270\tiny{±0.003} & 0.413\tiny{±0.004} & 0.503\tiny{±0.004} \\
\midrule
Ours & \textbf{0.602\tiny{±0.003}} & \textbf{0.780\tiny{±0.003}} & \textbf{0.868\tiny{±0.003}} \\
\bottomrule
\end{tabularx}
\label{tab:results_dg}
\end{table}

\begin{table*}[t]
\centering
\scriptsize
\newcolumntype{C}{>{\centering\arraybackslash}X}
\caption{Ablation study on MIMIC-CXR and IU X-Ray.}
\begin{tabularx}{\linewidth}{llCCCCCClCCCl}
\toprule	 
\multirow{2}{*}{\textbf{Dataset}} & \multirow{2}{*}{\textbf{Model}} & \multicolumn{7}{c}{\textbf{CE Metrics}} & \multicolumn{4}{c}{\textbf{NLG Metrics}}\\ \cmidrule(r){3-9} \cmidrule(r){10-13}
 & &Loc\textsubscript{h} & Loc\textsubscript{v} & Sev & F1 & Rec & Pre & Avg.& B1 & B4 & R & Avg.\\
\midrule
\multirow{4}{*}{\textbf{MIMIC}} & Baseline & 0.433 & 0.453 & 0.719 & 0.362 & 0.340 & 0.446 & 0.459 & 0.413 & 0.128 & 0.288 & 0.276\\
& +Omni & 0.436 & 0.456 & 0.719 & 0.362 & 0.342 & 0.445 & 0.460 (+0.2\%) & 0.413 & 0.127 & 0.286 & 0.275 (-0.4\%) \\
& +Omni +CC & 0.442 & 0.458 & 0.723  & 0.370 & 0.346  & 0.452 & 0.465 (+1.3\%) & 0.416 & 0.131 & 0.287 & 0.278 (+0.7\%) \\
& +Omni +CC +Prompt & 0.514 & 0.495 & 0.749  & 0.479 & 0.521  & 0.487 & 0.541 (+17.9\%) & 0.416 & 0.129 & 0.286 & 0.277 (+0.4\%) \\
\midrule
\multirow{4}{*}{\textbf{IU X-Ray}} & Baseline & 0.345 & 0.460 & 0.691  & 0.158 & 0.156  & 0.171 & 0.330 & 0.381 & 0.087 & 0.283 & 0.250 \\
& +Omni & 0.351 & 0.459 & 0.695  & 0.152  & 0.149  & 0.157 & 0.327 (-0.9\%) & 0.379 & 0.085 & 0.286 & 0.250 (+0.0\%) \\
& +Omni +CC & 0.363 & 0.474 & 0.709  & 0.178  & 0.177  & 0.194 & 0.349 (+5.8\%) & 0.403 & 0.093 & 0.286 & 0.261 (+4.4\%) \\
& +Omni +CC +Prompt & 0.404 & 0.514 & 0.732  & 0.219  & 0.234  & 0.218 & 0.387 (+17.3\%) & 0.403 & 0.091 & 0.288 & 0.261 (+4.4\%) \\
\bottomrule
\end{tabularx}
\label{tab:ablation}
\end{table*}

\begin{table}[t]
\centering
\scriptsize
\newcolumntype{C}{>{\centering\arraybackslash}X}
\caption{Comparison of different QA generation designing on MIMIC-CXR. The best results are in \textbf{bold}.}
\begin{tabularx}{\linewidth}{CCCCCC}
\toprule	 
\textbf{Self-Q.} & \textbf{QA-Cont.} & Loc\textsubscript{h} & Loc\textsubscript{v} & Sev & Average \\ 
\midrule
\xmark & \xmark & 0.505 & 0.456 & 0.735 & 0.565\\
\xmark & \cmark & 0.491 & \textbf{0.497} & 0.724 & 0.571\\
\cmark & \xmark & 0.511 & 0.478 & 0.739 & 0.576\\
\cmark & \cmark & \textbf{0.514} & 0.495 & \textbf{0.749} & \textbf{0.586}\\
\bottomrule
\end{tabularx}
\label{tab:results_qa}
\end{table}

\noindent
\textbf{Implementation details.}
We use ImageNet pre-trained ResNet-101~\cite{HZR16} as the image encoder and Bert~\cite{DCL19} as the text encoder. We adopt Llama-2~\cite{TMS23} as the text decoder. LoRA~\cite{HWA22} is applied to the text decoder for efficient fine-tuning while the encoders are fully trainable. The rank and $\alpha$ of LoRA are both set to 32. The threshold is set to 0.7 and 0.9 for pseudo-labeling and diagnosis grounding, respectively. The temperature $\tau$ is 0.07. AdamW~\cite{LoH17} is used as the optimizer with a weight decay of 0.05. The initial learning rate is set to 3e-5 with a cosine learning rate schedule. There are 15 training epochs in total. The batch size is 12 and image size is 224. The model was implemented with PyTorch 2.1 and trained with one H800 GPU for about 56 hours.

\subsection{Results} 

\noindent
\textbf{Report generation.}
We compare our model with SOTA RRG methods, including R2Gen~\cite{CSC20}, M2TR~\cite{NGF21}, MKSG~\cite{YWG22}, CliBert~\cite{YaP22}, CVT2Dis.~\cite{NDK22}, M2KT~\cite{YWG23}, METrans.~\cite{WLW23}, KiUT~\cite{HZZ23}, DCL~\cite{LLC23}, RGRG~\cite{TMK23}, R2GenGPT~\cite{WLW23b}, PromptMRG~\cite{JCL24}, and I3+C2FD~\cite{LTC24}. Moreover, we have added generalist foundation models for comparison, including Med-Flamingo~\cite{MHW23}, RadFM~\cite{WZZ23}, InternVL~\cite{CWW24}, LLaVA-Med~\cite{LWZ24}, and MedDr~\cite{HNW24}, which are trained to tackle various medical tasks rather than RRG alone. Table~\ref{tab:results_sota} shows the results on both IU X-Ray and MIMIC-CXR. First of all, we can see from the table that the proposed CoD obtains the best performance on most of the entries. To be specific, CoD obtains the best results on 11 out of 12 entries within CE metrics, demonstrating its superiority in generating clinically accurate reports. Notably, our method outperforms the other methods considerably in attribute accuracy, thanks to the effective guidance of QA prompts. For example, our method achieves 0.514 and 0.404 of horizontal lesion location accuracy on MIMIC and IU, which have gained 19.3\% and 16.8\% improvement respectively over the second best method PromptMRG. In terms of NLG metrcis, CoD obtains best results on four out of six entries, indicating the superior capability of LLMs in generating fluent and relevant texts. Interestingly, the performance of generalist foundation models are far behind that of specialists, where MedDr performs the best among generalists. This is understandable as generalists are supposed to deal with a number of tasks as well as out-of-domain data, which inevitably degrades their competitiveness on a specific task/data.

\noindent
\textbf{Lesion grounding.}
We show the effectiveness of our lesion grounding module by comparing it with the baseline framework~\cite{DYC21} we adopted. Table~\ref{tab:results_vg} gives the results tested on RSNA and VinDR-CXR. As can be seen, the baseline model achieves an average accuracy of 0.336, which is trained on the private dataset with GT annotations. By applying a vanilla omni-supervised learning strategy, more data from MIMIC-CXR is added to the training, providing reliable pseudo-boxes. The accuracy is improved to 0.354, indicating that more training data would benefit the performance, even without GTs. When the clinical consistency is further utilized to produce more accurate pseudo boxes, the average accuracy is boosted to 0.401, with a relative improvement of 19.3\% compared to the baseline.

\noindent
\textbf{Diagnosis grounding.}
We implemented four baselines for comparison: 1) keyword matching, i.e., converting QA diagnosis/sentence into a vector by checking whether the predefined disease keywords are present; 2) Bert~\cite{DCL19}; 3) CLIP~\cite{RKH21}; 4) CXR-CLIP~\cite{EKK21}. The results are listed in Table~\ref{tab:results_dg}. We found that our Bert-based model outperforms the baselines by a large margin, obtaining top-1 accuracy of 0.602. In contrast, Bert and CLIP achieves 0.114 and 0.202 in top-1 accuracy, respectively. Surprisingly, keyword matching even performs slightly better than Bert, probably due to its domain knowledge imposed via keywords. The MIMIC pretrained CXR-CLIP obtains better results than that of general CLIP, but still being inferior to our model (0.270 vs 0.602 in Top-1 accuracy). The above results imply the importance of scenario-specific tuning, especially for tasks that require fine-grained semantics.

\subsection{Model Analysis}

\textbf{Ablation study.}
To verify the effectiveness of the proposed modules, we conduct ablation study on both MIMIC-CXR and IU X-Ray, the results of which are given in Table~\ref{tab:ablation}. As can be seen, the baseline model obtains decent performance on both datasets, which is a conventional encoder-decoder model using the same encoder and decoder as that of CoD and trained on MIMIC. We then equip the baseline with vanilla omni-supervised learning by adding the private data to the training set and generate pseudo-labels via confidence-based thresholding. We can see from the table (row 2) that this method obtains marginal improvement in the CE metrics of MIMIC while the performance degrades in the NLG metrics of MIMIC and in the CE metrics of IU. It indicates that a naive way of omni-supervised learning may not work well due to the inevitable noise from pseudo-labels. When the clinical consistency constraint is further imposed during pseudo-labeling, we observe a considerable improvement (row 3), especially on IU. For example, the average CE metrics has been improved by 5.8\% and the improvement on NLG is 4.4\%. This is understandable as the usage of the private data increases the generalization of the model, which would gain more improvement on cross-domain data. Finally, we apply all the modules to restore the proposed CoD and it outperforms the baseline consistently on both datasets (row 4). In particular, its improvement on CE metrics is significant, yielding 17.9\% and 17.3\% increase against the baseline on MIMIC and IU, respectively. 

\begin{figure*}[t]
\centering
  \includegraphics[width=1\linewidth]{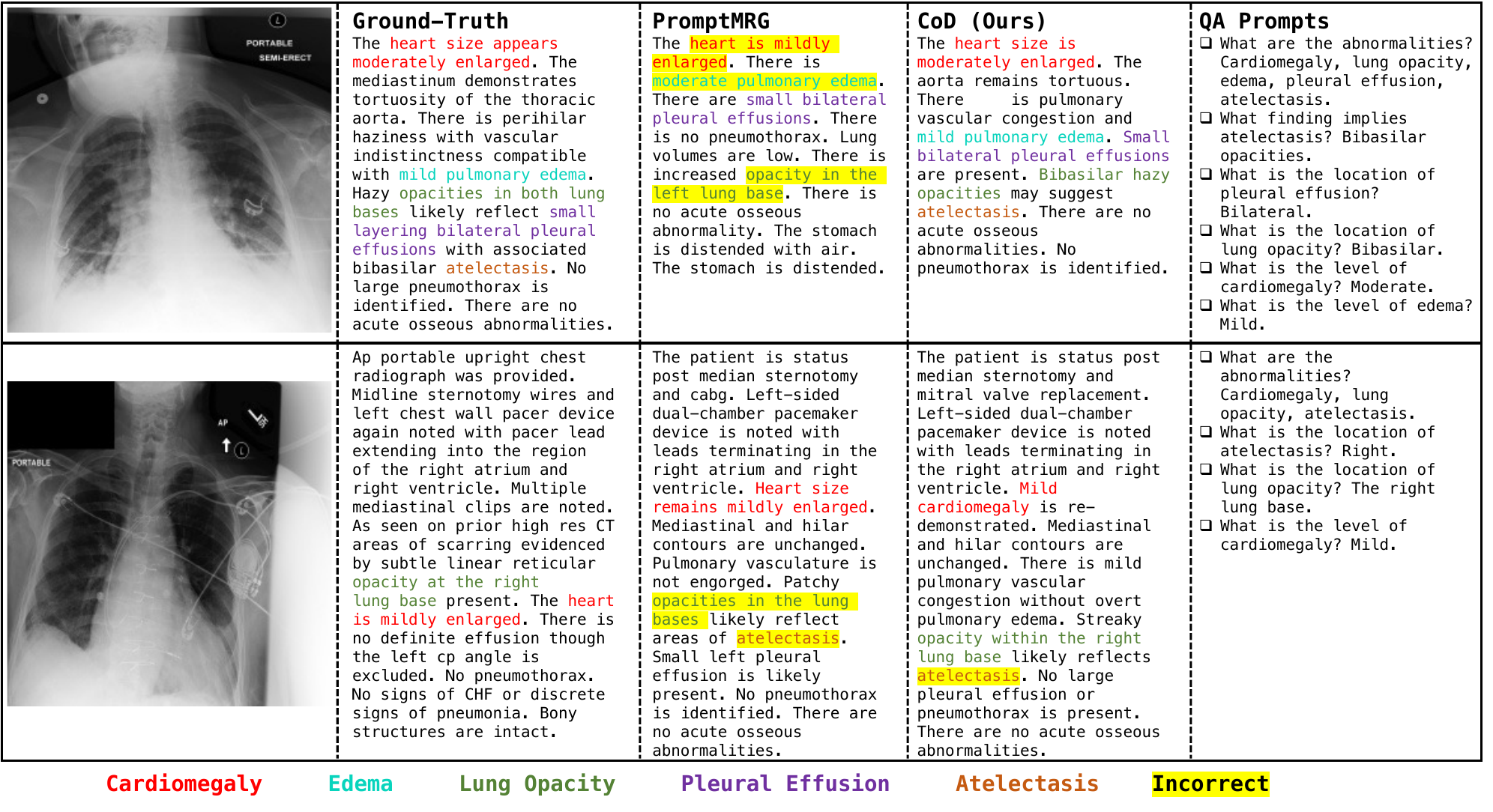}
\caption{Qualitative examples of PromptMRG~\cite{JCL24} and the proposed method CoD on report generation, selected from MIMIC-CXR test set. Different diseases are highlighted with different colors of fonts. Incorrect predictions are in yellow background. We also provide the intermediate QA prompts from CoD.} 
\label{fig:vis_compare}      
\end{figure*}

\begin{figure*}[t]
\centering
  \includegraphics[width=1\linewidth]{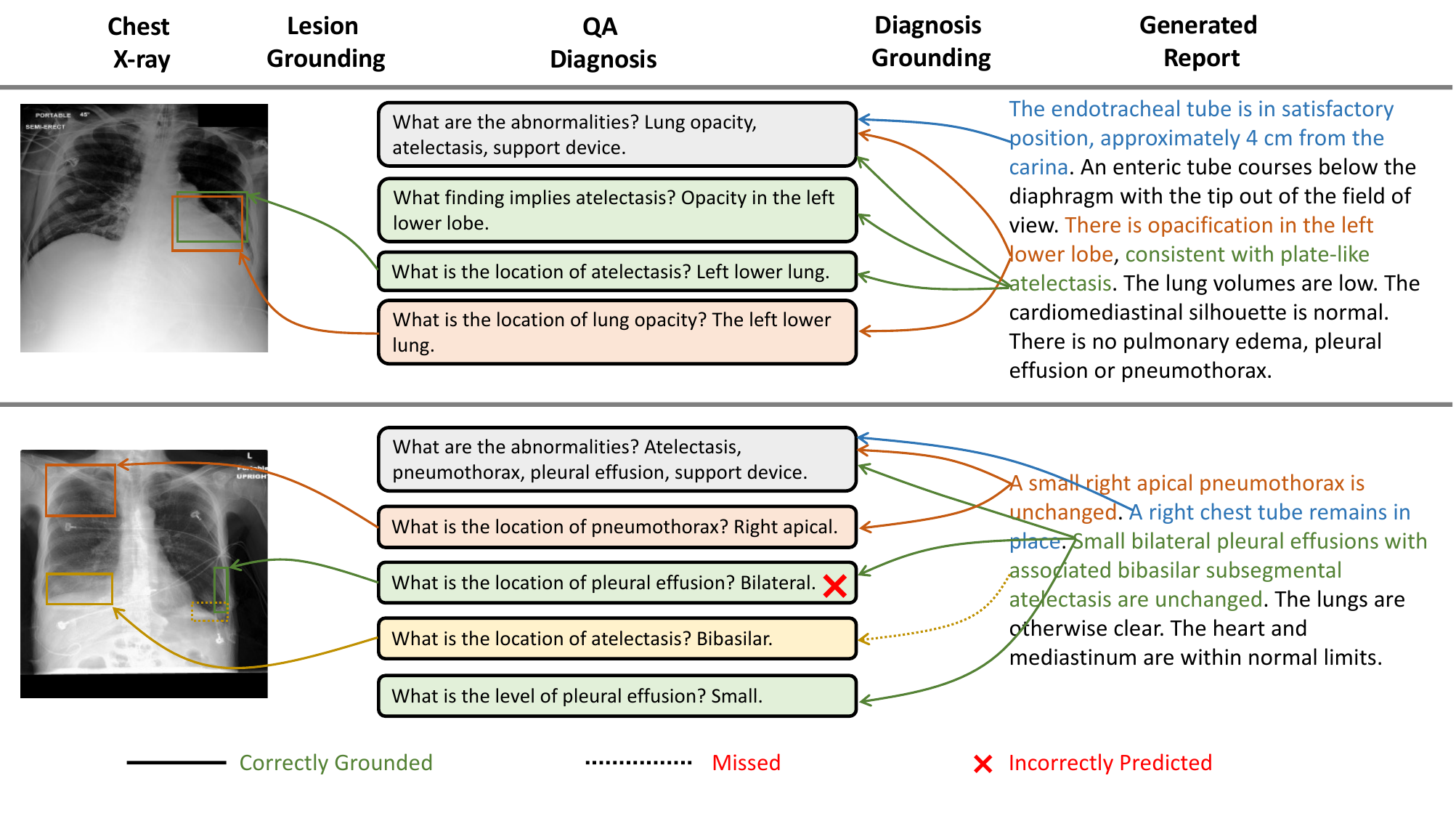}
\caption{Qualitative examples of the proposed method on diagnosis and lesion grounding, selected from MIMIC-CXR test set. Different diseases are highlighted with different colors. Solid lines are used for correctly grounded objects and dotted lines are for missed groundings. Wrong predictions of attributes in QA diagnosis are marked with cross.} 
\label{fig:vis_ground}      
\end{figure*}

\noindent
\textbf{QA generation designing.}
We investigate how the design of QA generation affects reasoning capability in attribute description. Our analysis identifies two critical design variables: (1) \textbf{Self-questioning}: Whether the model generates questions autonomously or follows predefined questions and (2) \textbf{QA-context}: Whether the model accesses previous QA pairs when answering the current question. While our proposed framework adopts both features, we systematically evaluate how performance varies across different design configurations. As shown in Table~\ref{tab:results_qa}, the configuration using predefined questions without QA-context (Row 1) yields the poorest attribute prediction performance. Introducing QA-context (Row 2) significantly improves vertical location accuracy ($+9.0\%$), demonstrating its importance for discriminating lesions across anatomical subregions. However, this comes with slight decreases in horizontal location ($-2.8\%$) and severity ($-1.5\%$) prediction accuracy. Using predefined questions could reduce the occasionally missed QAs from self-questions, but surprisingly, it does not necessarily improve accuracy. Through case analysis, we identified two phenomena explaining these results: 1) In most cases where the model "forgets" to ask about location/severity, the abnormalities were ultimately determined to be negative (false positives); 2) The LLM sometimes correctly infers and reports attributes even without explicit QA prompts about them. These findings suggest that while predefined questions could reduce omissions, they may also introduce unnecessary rigidity that slightly degrades overall performance. The autonomous questioning configuration without QA-context (Row 3) shows consistent improvements across all attributes compared to the baseline. Notably, the combination of both self-questioning and QA-context (Row 4) achieves optimal overall performance, suggesting that end-to-end optimization of QA generation is particularly effective for fine-grained attribute prediction.

\subsection{Qualitative Results}

\noindent
\textbf{Report generation.}
Two qualitative examples are given in Fig.~\ref{fig:vis_compare} to show the superiority of CoD over PromptMRG. We highlight each disease with a different color for convenience and the incorrect predictions are highlighted in yellow background. We can see that CoD not only predicts the diseases correctly, but also provides precise descriptions of the lesions. In contrast, PromptMRG performs well in predicting diagnostic results while its descriptions of lesions are unsatisfactory. For example, CoD successfully predicts the severity of cardiomegaly and edema for the first image while PromptMRG gives incorrect severities for both diseases; in the second image, CoD identifies the location of opacity should be at the right lung base while PromptMRG predicts it to be at both sides, which is again wrong. To understand how QA prompting helps improve the descriptive accuracy, we further show the intermediate QA prompts generated by our model. It can be observed that these prompts are mostly correct and well aligned with the generated reports, demonstrating the promise of this technique. However, it may miss some attributes occasionally, such as the severity of pleural effusion in the first image. We attribute this issue to the imperfect training data, where the intermediate prompts are not always complete. 

\noindent
\textbf{Grounding.}    
Fig.~\ref{fig:vis_ground} presents two qualitative examples demonstrating our diagnosis and lesion grounding results. In the first simpler case, the model correctly grounds both report sentences to their corresponding QA diagnoses and accurately links these to the relevant image regions. Note that support devices are excluded from grounding as they fall outside our detection scope. The second, more complex example shows our model successfully handling more abnormalities. (1) \textit{Pneumothorax}: Correctly predicted in terms of presence and location, with proper linkages between the report sentence, QA pair, and bounding box for rapid localization. (2) \textit{Pleural effusion}: Accurately identified, quantified, and localized, though the QA generation initially misassigned its location to both lungs (should be left lung only). Notably, while the QA generation missed pneumothorax severity and initially erred in pleural effusion localization, the final report and bounding boxes remained correct. This suggests our prompting and grounding modules exhibit error tolerance by not strictly depending on QA outputs. One limitation appears with \textit{atelectasis}, where grounding failed due to: (i) sentence-level disease mixing (with pleural effusion), and (ii) missed detection at the left lung base. These cases highlight challenges in handling co-occurring pathologies. Overall, these visualizations demonstrate how our grounding modules streamline radiologists' workflow by enabling efficient verification and correction.

\section{Conclusion}
In this work, we proposed a RRG framework, CoD, to tackle the challenges in clinical efficacy and explainability of report generation. Specifically, a diagnostic conversation module is proposed to first extract key findings from the image via QA pairs, then it prompts the LLM with these QA diagnoses for accurate report generation. The explainability is enhanced by connecting the generated sentences to the intermediate QA diagnoses such that our model is no longer a black box but a system with traceable clues. Moreover, the location-related QA diagnoses will be grounded to the image, further providing convenience for radiologists. Our extensive experiments on two RRG benchmarks, two lesion grounding datasets, and one diagnosis grounding data demonstrate the excellence of our model. 

As future work, we plan to extend the CoD framework in three key directions: (1) expanding input compatibility to 3D radiological data (e.g., CT and MRI scans) for broader clinical applicability, (2) enhancing QA diagnosis diversity through LLM-assisted question generation and improved data curation pipelines to ensure higher-quality training data, and (3) optimizing computational efficiency to maintain practical usability despite the increased processing demands of volumetric medical imaging. These improvements will address current limitations while preserving the framework's clinical utility and user experience.

\nocite{*} % Force all references to appear
\bibliographystyle{IEEEtran}
\bibliography{mybib}

\end{document}